%% file: acl2017.tex
\documentclass[11pt,a4paper]{article}
\usepackage[nohyperref]{acl2017}
\usepackage{times}
\usepackage{latexsym}
\usepackage{url}
\usepackage{multirow}
\usepackage{rotating}
\usepackage{amsmath}
\usepackage{graphicx}
\usepackage{color}
\usepackage[small]{caption}
\usepackage{placeins}
\usepackage{wrapfig}
\usepackage{sidecap}
\usepackage{booktabs}
\aclfinalcopy
\def\aclpaperid{3470}

\title{Differentiable Scheduled Sampling for Credit Assignment}
\author{Kartik Goyal \\
  Carnegie Mellon University\\
  Pittsbrugh, PA, USA\\
  {\tt kartikgo@cs.cmu.edu} \\\And
  Chris Dyer \\
  DeepMind \\
  London, UK \\
  {\tt cdyer@google.com} \\\And
  Taylor Berg-Kirkpatrick \\
  Carnegie Mellon University \\
  Pittsburgh, PA, USA \\
  {\tt tberg@cs.cmu.edu} \\}
\begin{document}
\maketitle
\begin{abstract}

We demonstrate that a continuous relaxation of the argmax operation can be used to create a differentiable approximation to greedy decoding for sequence-to-sequence (seq2seq) models. By incorporating this approximation into the scheduled sampling training procedure \cite{bengioss}--a well-known technique for correcting exposure bias--we introduce a new training objective that is continuous and differentiable everywhere and that can provide informative gradients near points where previous decoding decisions change their value. In addition, by using a related approximation, we demonstrate a similar approach to sampled-based training. Finally, we show that our approach outperforms cross-entropy training and scheduled sampling procedures in two sequence prediction tasks: named entity recognition and machine translation.

\end{abstract}

\section{Introduction}

Sequence-to-Sequence (seq2seq) models have demonstrated excellent performance in several tasks including machine translation \cite{sutskever2014sequence}, summarization \cite{rush2015neural}, dialogue generation \cite{serban2015building}, and image captioning \cite{xu2015show}. 
However, the standard cross-entropy training procedure for these models suffers from the well-known problem of exposure bias:  because cross-entropy training always uses gold contexts, the states and contexts encountered during training do not match those encountered at test time. This issue has been addressed using several approaches that try to incorporate awareness of decoding choices into the training optimization. These include reinforcement learning \cite{ranzato2015sequence, bahdanau2016actor}, imitation learning \cite{daume2009search, ross2011reduction, bengioss}, and beam-based approaches \cite{wiseman2016sequence, andor2016globally, daume2005learning}. In this paper, we focus on one the simplest to implement and least computationally expensive approaches, \textit{scheduled sampling} \cite{bengioss}, which stochastically incorporates contexts from previous decoding decisions into training. 

While scheduled sampling has been empirically successful, its training objective has a drawback: because the procedure directly incorporates greedy decisions at each time step, the objective is discontinuous at parameter settings where previous decisions change their value. As a result, gradients near these points are non-informative and scheduled sampling has difficulty assigning credit for errors. In particular, the gradient does not provide information useful in distinguishing between local errors without future consequences and cascading errors which are more serious.


Here, we propose a novel approach based on scheduled sampling that uses a differentiable approximation of previous greedy decoding decisions inside the training objective by incorporating a continuous relaxation of argmax. As a result, our end-to-end relaxed greedy training objective is differentiable everywhere and fully continuous. By making the objective continuous at points where previous decisions change value, our approach provides gradients that can respond to cascading errors. In addition, we demonstrate a related approximation and reparametrization for sample-based training (another training scenario considered by scheduled sampling \cite{bengioss}) that can yield stochastic gradients with lower variance than in standard scheduled sampling. In our experiments on two different tasks, machine translation (MT) and named entity recognition (NER), we show that our approach outperforms both cross-entropy training and standard scheduled sampling procedures with greedy and sampled-based training.

\section{Discontinuity in Scheduled Sampling}
 
While scheduled sampling \cite{bengioss} is an effective way to rectify exposure bias, it cannot differentiate between cascading errors, which can lead to a sequence of bad decisions, and local errors, which have more benign effects. Specifically, scheduled sampling focuses on learning optimal behavior in the current step given the fixed decoding decision of the previous step. If a previous bad decision is largely responsible for the current error, the training procedure has difficulty adjusting the parameters accordingly. The following machine translation example highlights this credit assignment issue:\\[0.2cm]
\begin{centering}\includegraphics[width=0.43\textwidth]{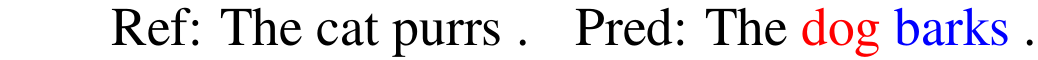}\end{centering}\\[0.2cm]
\noindent At step 3, the model prefers the word `barks' after incorrectly predicting `dog' at step 2. To correct this error, the scheduled sampling procedure would increase the score of `purrs' at step 3, conditioned on the fact that the model predicted (incorrectly) `dog' at step 2, which is not the ideal learning behaviour. Ideally, the model should be able to backpropagate the error from step 3 to the source of the problem which occurred at step 2, where `dog' was predicted instead of `cat'. 



\begin{figure}[t]
\centering
\includegraphics[height=4.3cm]{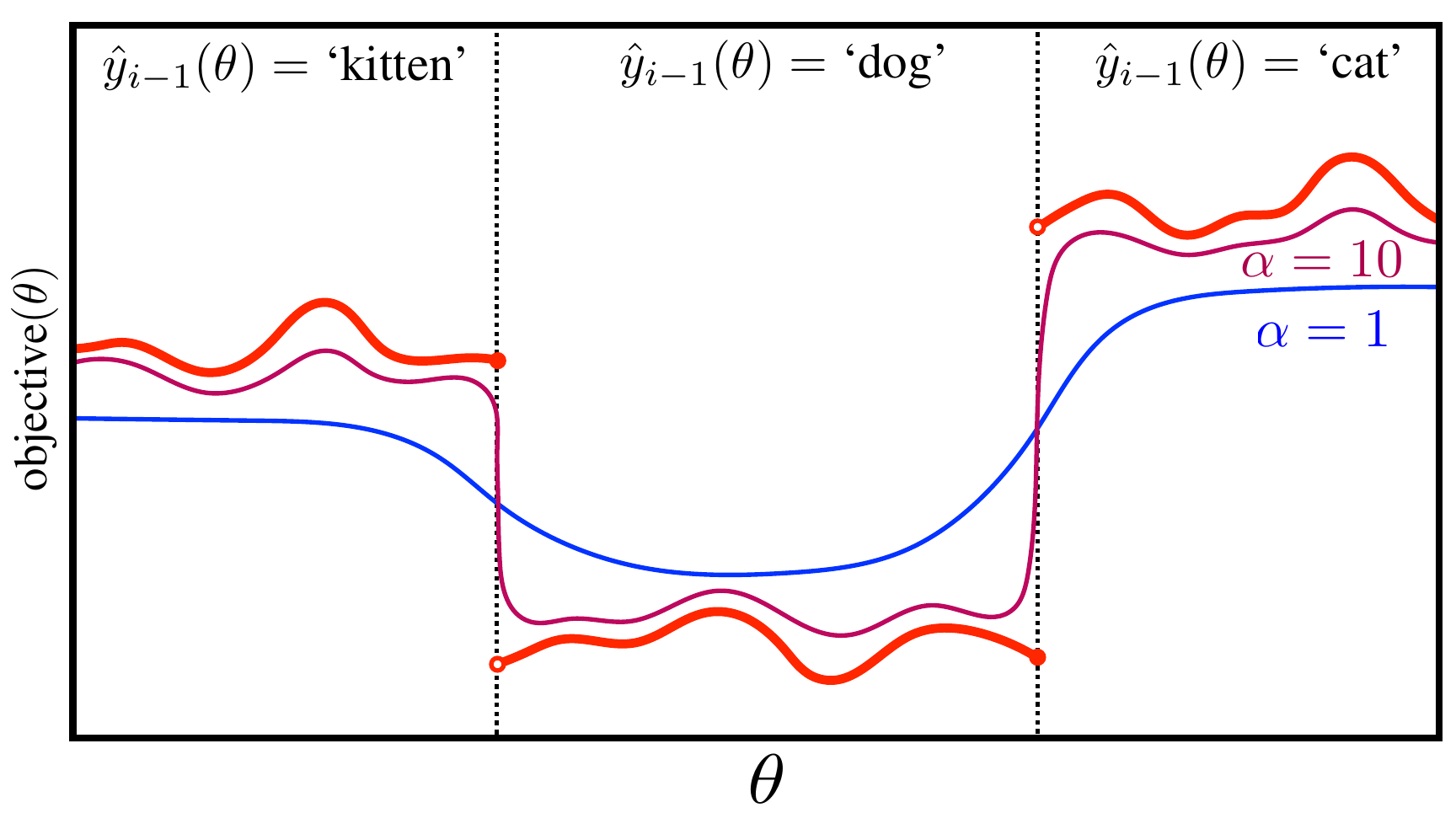}
  \caption{\label{discont_fig} Discontinuous scheduled sampling objective (red) and continuous relaxations (blue and purple).}
\end{figure}

The lack of credit assignment during training is a result of discontinuity in the objective function used by scheduled sampling, as illustrated in Figure~\ref{discont_fig}. We denote the ground truth target symbol at step $i$ by $y^*_i$, the embedding representation of word $y$ by $e(y)$, and the hidden state of a seq2seq decoder at step $i$ as $h_i$. Standard cross-entropy training defines the loss at each step to be $\log p(y^*_i | h_i(e(y^{*}_{i-1}), h_{i-1}))$, while scheduled sampling uses loss $\log p(y^*_i | h_i(e(\hat{y}_{i-1}), h_{i-1}))$, where $\hat{y}_{i-1}$ refers the model's prediction at the previous step.\footnote{For the sake of simplicity, the `always sample' variant of scheduled sampling is described \cite{bengioss}.} Here, the model prediction $\hat{y}_{i-1}$ is obtained by argmaxing over the output softmax layer. 
Hence, in addition to the intermediate hidden states and final softmax scores, the previous model prediction, $\hat{y}_{i-1}$, itself depends on the model parameters, $\theta$, and ideally, should be backpropagated through, unlike the gold target symbol $y^{*}_{i-1}$ which is independent of model parameters. 
However, the argmax operation is discontinuous, and thus the training objective (depicted in Figure~\ref{discont_fig} as the red line) exhibits discontinuities at parameter settings where the previous decoding decisions change value (depicted as changes from `kitten' to `dog' to `cat'). Because these change points represent discontinuities, their gradients are undefined and the effect of correcting an earlier mistake (for example `dog' to `cat') as the training procedure approaches such a point is essentially hidden.

In our approach, described in detail in the next section, we attempt to fix this problem by incorporating a continuous relaxation of the argmax operation into the scheduled sampling procedure in order to form an approximate but fully continuous objective. Our relaxed approximate objective is depicted in Figure~\ref{discont_fig} as blue and purple lines, depending on temperature parameter $\alpha$ which trades-off smoothness and quality of approximation.  

\begin{figure}[t]
\centering
\includegraphics[height=4.8cm]{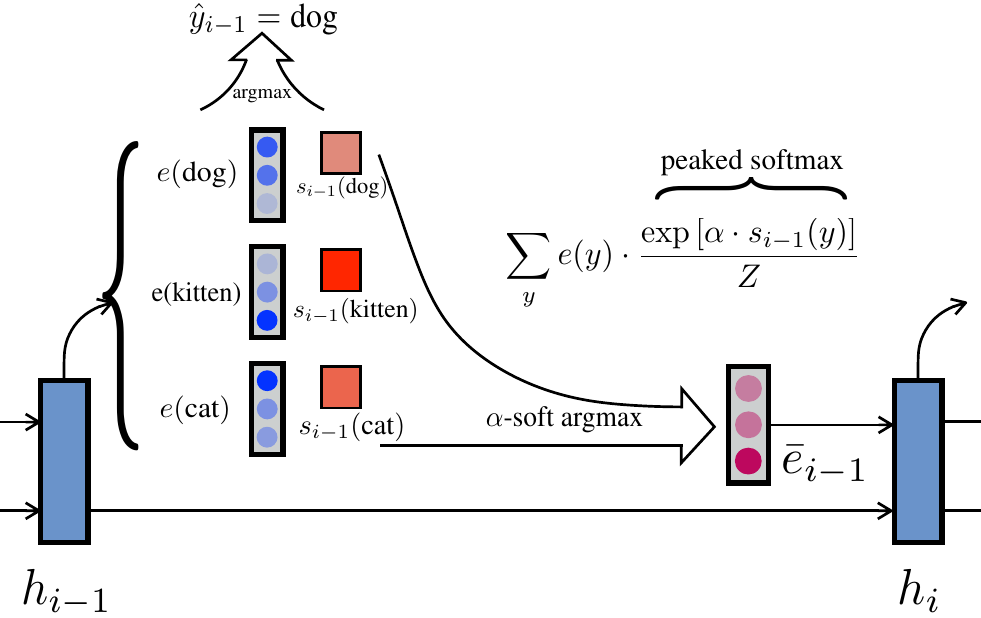}
  \caption{\label{decoder_fig} Relaxed greedy decoder that uses a continuous approximation of argmax as input to the decoder state at next time step.}
\end{figure}

\section{Credit Assignment via Relaxation} 

In this section we explain in detail the continuous relaxation of greedy decoding that we will use to build a fully continuous training objective. We also introduce a related approach for sample-based training. 

\subsection{Soft Argmax \label{softargmax}}

In scheduled sampling, the embedding for the best scoring word at the previous step is passed as an input to the current step. This operation\footnote{Assuming there are no ties for the sake of simplicity.} can be expressed as 
\begin{align*}
    \hat{e}_{i-1} = \sum_y e(y)\boldsymbol{1}[\forall y' \ne y\ \ s_{i-1}(y) > s_{i-1}(y')]
\end{align*}
where $y$ is a word in the vocabulary, $s_{i-1}(y)$ is the output score of that word at the previous step, and $\hat{e}_{i-1}$ is the embedding passed to the next step.
This operation can be relaxed by replacing the indicator function with a \emph{peaked softmax function} with hyperparameter $\alpha$ to define a soft argmax procedure:
\begin{align*}
    \bar{e}_{i-1} = \sum_y e(y) \cdot \frac{\exp{(\alpha~ s_{i-1}(y)})}{\sum_{y'} \exp{(\alpha~ s_{i-1}(y'))}}
\end{align*}
As $\alpha \to \infty$, the equation above approaches the true argmax embedding. Hence, with a finite and large $\alpha$, we get a linear combination of all the words (and therefore a continuous function of the parameters) that is dominated heavily by the word with maximum score.

\subsection{Soft Reparametrized Sampling \label{concrete}}

Another variant of scheduled sampling is to pass a sampled embedding from the softmax distribution at the previous step to the current step instead of the argmax. This is expected to enable better exploration of the search space during optimization due to the added randomness and hence result in a more robust model. In this section, we discuss and review an approximation to the Gumbel reparametrization trick that we use as a module in our sample-based decoder.
This approximation was proposed by \newcite{maddison2016concrete} and \newcite{jang2016categorical}, who showed that the same soft argmax operation introduced above can be used 
for reducing variance of stochastic gradients when sampling from softmax distributions. Unlike soft argmax, this approach is not a fully continuous approximation to the sampling operation, but it does result in much more informative gradients compared to naive scheduled sampling procedure.

The Gumbel reparametrization trick shows that sampling from a categorical distribution can be refactored into sampling from a simple distribution followed by a deterministic transformation as follows: (i) sampling an independent Gumbel noise $G$ for each element in the categorical distribution, typically done by transforming a sample from the uniform distribution: $U \sim Uniform(0,1)$ as $G = -log(-log~U)$, then (ii) adding it componentwise to the unnormalized score of each element, and finally (iii) taking an argmax over the vector. Using the same argmax softening procedure as above, they arrive at an approximation to the reparametrization trick which mitigates some of the gradient's variance introduced by sampling. The approximation is\footnote{This is different from using the expected softmax embedding because our approach approximates the actual sampling process instead of linearly weighting the embeddings by their softmax probabilities}:
\begin{align*}
    \tilde{e}_{i-1} = \sum_{y} e(y) \cdot
    \frac{\exp{(\alpha~(s_{i-1}(y) + G_y))}}{ \sum_{y'} \exp{(\alpha~(s_{i-1}(y') + G_{y'}))}}
\end{align*}
We will use this `concrete' approximation of softmax sampling in our relaxation of scheduled sampling with a sample-based decoder. We discuss details in the next section. Note that our original motivation based on removing discontinuity does not strictly apply to this sampling procedure, which still yields a stochastic gradient due to sampling from the Gumbel distribution. However, this approach is conceptually related to greedy relaxations since, here, the soft argmax reparametrization reduces gradient variance which may yield a more informative training signal. Intuitively, this approach results in the gradient of the loss to be more aware of the sampling procedure compared to naive scheduled sampling and hence carries forward information about decisions made at previous steps. The empirical results, discussed later, show similar gains to the greedy scenario.

\subsection{Differentiable Relaxed Decoders}

With the argmax relaxation introduced above, we have a recipe for a fully differentiable greedy decoder designed to produce informative gradients near change points. Our final training network for scheduled sampling with relaxed greedy decoding is shown in Figure~\ref{decoder_fig}. Instead of conditioning the current hidden state, $h_i$, on the argmax embedding from the previous step, $\hat e_{i-1}$, we use the $\alpha$-soft argmax embedding, $\bar e_{i-1}$, defined in Section~\ref{softargmax}. This removes the discontinuity in the original greedy scheduled sampling objective by passing a linear combination of embeddings, dominated by the argmax, to the next step. Figure~\ref{discont_fig} illustrates the effect of varying $\alpha$. As $\alpha$ increases, we more closely approximate the greedy decoder.

As in standard scheduled sampling, here we minimize the cross-entropy based loss at each time step. Hence the computational complexity of our approach is comparable to standard seq2seq training. As we discuss in Section~\ref{expers}, mixing model predictions randomly with ground truth symbols during training \cite{bengioss, daume2009search, ross2011reduction}, while annealing the probability of using the ground truth with each epoch, results in better models and more stable training. As a result, training is reliant on the \emph{annealing schedule} of two important hyperparameters: i) ground truth mixing probability and ii) the \emph{$\alpha$ parameter} used for approximating the argmax function. For output prediction, at each time step, we can still output the hard argmax, depicted in Figure~\ref{decoder_fig}.  

For the case of scheduled sampling with sample-based training--where decisions are sampled rather than chosen greedily \cite{bengioss}--we conduct experiments using a related training procedure. Instead of using soft argmax, we use the soft sample embedding, $\tilde e_{i-1}$, defined in Section~\ref{concrete}. Apart from this difference, training is carried out using the same procedure. 
\input{table.tex}
\section{Related Work}
\newcite{gormley2015approximation}'s approximation-aware training is conceptually related, but focuses on variational decoding procedures. \newcite{hoang2017decoding} also propose continuous relaxations of decoders, but are focused on developing better inference procedures.
\newcite{grefenstette2015learning} successfully use a soft approximation to argmax in neural stack mechanisms. Finally, \newcite{ranzato2015sequence} experiment with a similarly motivated objective that was not fully continuous, but found it performed worse than the standard training.

\section{Experimental Setup \label{expers}}
 We perform experiments with machine translation (MT) and named entity recognition (NER).\\[0.15cm]
 \noindent {\bf Data:} For MT, we use the same dataset (the German-English portion of the IWSLT 2014 machine translation evaluation campaign \cite{cettolo2014report}), preprocessing and data splits as \newcite{ranzato2015sequence}. For named entity recognition, we use the CONLL 2003 shared task data \cite{tjong2003introduction} for German language and use the provided data splits. We perform no preprocessing on the data.The output vocabulary length for MT is 32000 and 10 for NER.\\[0.15cm]
 \noindent {\bf Implementation details:} For MT, we use a seq2seq model with a simple attention mechanism \cite{bahdanau2014neural}, a bidirectional LSTM encoder (1 layer, 256 units), and an LSTM decoder (1 layer, 256 units). For NER, we use a seq2seq model with an LSTM encoder (1 layer, 64 units) and an LSTM decoder (1 layer, 64 units) with a fixed attention mechanism that deterministically attends to the $i$th input token when decoding the $i$th output, and hence does not involve learning of attention parameters. 
\footnote{\emph{Fixed attention} refers to the scenario when we use the bidirectional LSTM encoder representation of the source sequence token at time step $t$ while decoding at time step $t$ instead of using a linear combination of all the input sequences weighted according to the attention parameters in the standard attention mechanism based models.}
 \\[0.15cm]
\noindent {\bf Hyperparameter tuning:} We start by training with actual ground truth sequences for the first epoch and decay the probability of selecting the ground truth token as an inverse sigmoid \cite{bengioss} of epochs with a decay strength parameter $k$. We also tuned for different values of $\alpha$ and explore the effect of varying $\alpha$ exponentially (annealing) with the epochs. In table~\ref{results}, we report results for the best performing configuration of decay parameter and the $\alpha$ parameter on the validation set. To account for variance across randomly started runs, we ran multiple random restarts (RR) for all the systems evaluated and always used the RR with the best validation set score to calculate test performance.\\[0.15cm]
\noindent {\bf Comparison} We report validation and test metrics for NER and MT tasks in Table~\ref{results}, F1 and BLEU respectively. `Greedy' in the table refers to scheduled sampling with soft argmax decisions (either soft or hard) and `Sample' refers the corresponding reparametrized sample-based decoding scenario. We compare our approach with two baselines: standard cross-entropy loss minimization for seq2seq models (`Baseline CE') and the standard scheduled sampling procedure (\newcite{bengioss}). We report results for two variants of our approach: one with a fixed $\alpha$ parameter throughout the training procedure ($\alpha$-soft fixed), and the other in which  we vary $\alpha$ exponentially with the number of epochs ($\alpha$-soft annealed).
\section{Results}
All three approaches improve over the standard cross-entropy based seq2seq training. Moreover, both approaches using continuous relaxations (greedy and sample-based) outperform standard scheduled sampling \cite{bengioss}. The best results for NER were obtained with the relaxed greedy decoder with annealed $\alpha$ which yielded an F1 gain of +3.1 over the standard seq2seq baseline and a gain of +1.5 F1 over standard scheduled sampling. For MT, we obtain the best results with the relaxed sample-based decoder, which yielded a gain of +1.5 BLEU over standard seq2seq and a gain of +0.75 BLEU over standard scheduled sampling. 

We observe that the reparametrized sample-based method, although not fully continuous end-to-end unlike the soft greedy approach, results in good performance on both the tasks, particularly MT. This might be an effect of stochastic exploration of the search space over the output sequences during training and hence we expect MT to benefit from sampling due to a much larger search space associated with it. 
\input{table2.tex}
We also observe that annealing $\alpha$ results in good performance which suggests that a smoother approximation to the loss function in the initial stages of training is helpful in guiding the learning in the right direction. However, in our experiments we noticed that the performance while annealing $\alpha$ was sensitive to the hyperparameter associated with the annealing schedule of the mixing probability in scheduled sampling during training.

The computational complexity of our approach is comparable to that of standard seq2seq training. However, instead of a vocabulary-sized max and lookup, our approach requires a matrix multiplication. Practically, we observed that on GPU hardware, all the models for both the tasks had similar speeds which suggests that our approach leads to accuracy gains without compromising run-time. 
Moreover, as shown in Table~\ref{results2}, we observe that a gradual decay of mixing probability consistently compared favorably to more aggressive decay schedules. We also observed that the `always sample' case of relaxed greedy decoding, in which we never mix in ground truth inputs (see \newcite{bengioss}), worked well for NER but resulted in unstable training for MT. We reckon that this is an effect of large difference between the search space associated with NER and MT.
\section{Conclusion}
Our positive results indicate that mechanisms for credit assignment can be useful when added to the models that aim to ameliorate exposure bias. Further, our results suggest that continuous relaxations of the argmax operation can be used as effective approximations to hard decoding during training.



\section*{Acknowledgements}
We thank Graham Neubig for helpful discussions. We also thank the three anonymous reviewers for their valuable feedback.

\bibliographystyle{acl_natbib}

\bibliography{refs}

\end{document}

%% file: table.tex
\hspace{-0.2cm}\begin{table*}[t]
\centering
\caption{Result on NER and MT. We compare our approach ($\alpha$-soft argmax with fixed and annealed temperature) with standard cross entropy training (Baseline CE) and discontinuous scheduled sampling (\newcite{bengioss}). `Greedy' and `Sample' refer to Section~\ref{softargmax} and Section~\ref{concrete}.}
\label{results}
\scalebox{1.0}{
\begin{tabular}{lllllllll}
\toprule
Training procedure
& \multicolumn{4}{c}{NER (F1)} & \multicolumn{4}{c}{MT (BLEU)}\\
\midrule
& \multicolumn{2}{c}{Dev} & \multicolumn{2}{c}{Test} & \multicolumn{2}{c}{Dev} & \multicolumn{2}{c}{Test} \\
Baseline CE     &   \multicolumn{2}{c}{49.43}       & \multicolumn{2}{c}{53.32}   &     \multicolumn{2}{c}{20.35}        & \multicolumn{2}{c}{19.11} \\
\midrule
\multicolumn{1}{c}{} & \multicolumn{2}{c}{Greedy} & \multicolumn{2}{c}{Sample} & \multicolumn{2}{c}{Greedy} & \multicolumn{2}{c}{Sample} \\
\multicolumn{1}{c}{} & Dev          & Test        & Dev          & Test        & Dev         & Test         & Dev        & Test          \\
\midrule
\newcite{bengioss}   & 49.75        & 54.83       & 50.90        & 54.60       &    20.52     & 19.85        &     20.40       & 19.69        \\
$\alpha$-soft fixed  & 51.65        & 55.88       & 51.13        & 56.25       &         21.32    & 20.28        &    20.48        & 19.69        \\
$\alpha$-soft annealed   & 51.43        & \bf 56.33       & 50.99        & 54.20       &     21.28        & 20.18        &    21.36        & \bf 20.60 \\
\bottomrule
\end{tabular}
}
\end{table*}


%% file: table2.tex
\begin{table}[b!]
\centering
\noindent\scalebox{1.0}{
\begin{tabular}{|c|| c| c| c| c|} 
 \hline
 k & 100 & 10 & 1 & \textit{Always} \\
 \hline
 NER (F1) & 56.33 & 55.88 & 55.30 & 54.83\\
 \hline
\end{tabular}
}
\caption{Effect of different schedules for scheduled sampling on NER. k is the decay strength parameter. Higher k corresponds to gentler decay schedules. \textit{Always} refers to the case when predictions at the previous predictions are always passed on as inputs to the next step. }
\label{results2}
\end{table}